\documentclass[letterpaper, 10 pt, conference]{ieeeconf}  
\IEEEoverridecommandlockouts                              
\usepackage{amsmath,amsfonts,amssymb} 
\usepackage{algorithmic}
\usepackage{array}
\usepackage{url}
\usepackage{verbatim}
\def\BibTeX{{\rm B\kern-.05em{\sc i\kern-.025em b}\kern-.08em
		T\kern-.1667em\lower.7ex\hbox{E}\kern-.125emX}}
\usepackage{graphicx}
\usepackage{siunitx}
\usepackage{nolbreaks}
\usepackage{booktabs}
\usepackage[dvipsnames]{xcolor}
\usepackage{lipsum}
\usepackage[nolist,nohyperlinks]{acronym}
\usepackage[noadjust]{cite}
\usepackage{placeins}

\usepackage{subcaption}
\usepackage[ruled,vlined]{algorithm2e}

\acrodef{PSO}[PSO]{Particle Swarm Optimization}

\begin{document}

	\title{ \bf \nolbreaks{Quadratic Surrogate Attractor for Particle Swarm Optimization} } 
	\author{Maurizio Clemente$^{1}$ and Marcello Canova$^{1}$
		\thanks{$^{1}$Center for Automotive Research, The Ohio State University, Columbus, OH 43212 USA
			{\tt\small (email: clemente.52@osu.edu)}}}%
	\maketitle

	\begin{abstract} 
	This paper presents a particle swarm optimization algorithm that leverages surrogate modeling to replace the conventional global best solution with the minimum of an \textit{n}-dimensional quadratic form, providing a better‑conditioned dynamic attractor for the swarm.
	This refined convergence target, informed by the local landscape, enhances global convergence behavior and increases robustness against premature convergence and noise, while incurring only minimal computational overhead.
	The surrogate‑augmented approach is evaluated against the standard algorithm through a numerical study on a set of benchmark optimization functions that exhibit diverse landscapes. 
	To ensure statistical significance, 400 independent runs are conducted for each function and algorithm, and the results are analyzed based on their statistical characteristics and corresponding distributions.
	The quadratic surrogate attractor consistently outperforms the conventional algorithm across all tested functions.
	The improvement is particularly pronounced for quasi‑convex functions, where the surrogate model can exploit the underlying convex‑like structure of the landscape.

	\end{abstract}

\section{Introduction}\label{sec:introduction}

\begin{figure}[t]
	\centering
	\includegraphics[width=\columnwidth]{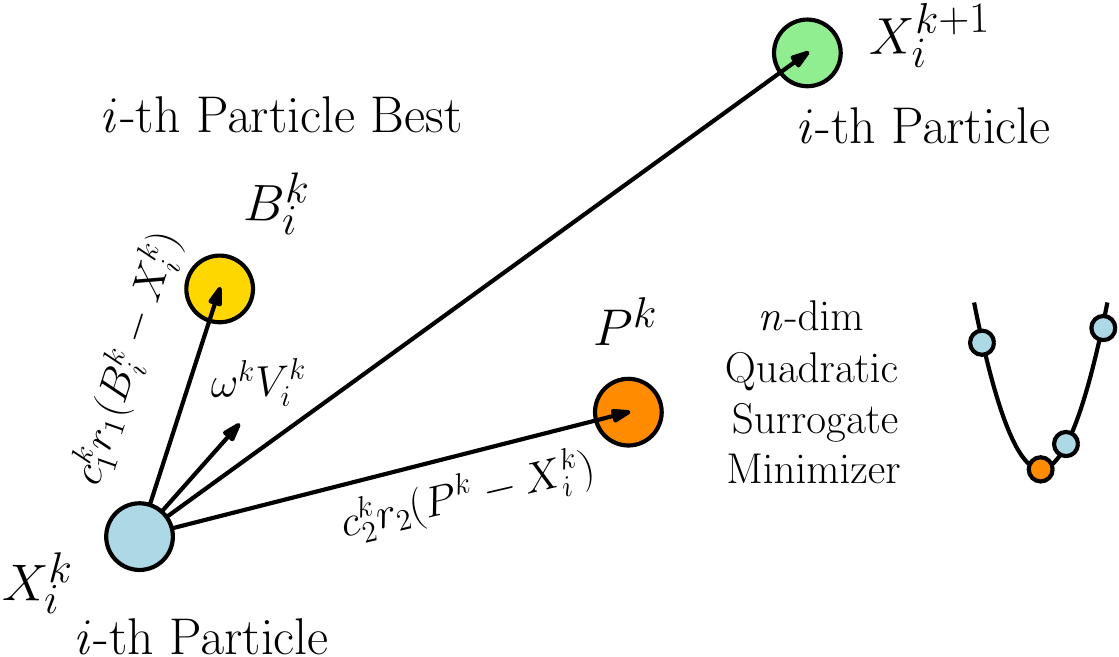}
	\caption{Schematic layout of the particle position $X_i^k$ update algorithm. In the standard approach, particles move following their speed $V_i^k$ and inertia $\omega$, the location of their own best evaluation $B_i^k$, and the overall best solution found. In the proposed method, the latter contribution is replaced by the minimum of a quadratic surrogate model $P^k$ constructed from multiple distinct best-performing locations.\vspace{-0.7cm}} 
\label{fig:Header}
\end{figure}

\ac{PSO} algorithms are widely recognized as effective tools for solving general optimization problems across a broad range of fields~\cite{Poli2008}.
Their combined flexibility and effectiveness stems from the use of simple stigmergic collaboration rules rather than gradient information to converge toward an optimum~\cite{KennedyEberhart1995}, which allows them to perform well even when the objective function is highly complex, non-differentiable, or non-smooth.
In line with other nature-inspired meta-heuristics, swarm algorithms rely heavily on numerous function evaluations to thoroughly explore the search domain.
However, most of the information generated from the many function evaluations is typically discarded, retaining only each particle’s personal best and the global best.
Rather than discarding knowledge acquired across the domain, it is used to construct a simple surrogate model of the function, improving the algorithm’s accuracy.
This approach is more resilient to premature convergence to local optima, as the surrogate function is constructed from multiple distinct best-performing locations, rather than being guided solely by a single global best, as in conventional algorithms.
These properties are particularly valuable when addressing highly complex multi‑objective problems, such as the integrated design optimization of lithium‑ion batteries, where coupled thermo‑mechanical and electro-chemical phenomena must be considered simultaneously. 
When analyzing trade‑offs between energy density and performance, the governing equations of high‑fidelity models are strongly nonlinear and tightly coupled, creating a search space riddled with local minima, while the evaluation of each candidate solution remains computationally expensive.
Consequently, the quadratic surrogate attractor approach is especially well suited to such problems, where accuracy, efficiency, and robustness in navigating intricate landscapes are essential.
Against this backdrop, this paper introduces the \ac{PSO} algorithm schematically represented in Fig.~\ref{fig:Header}, leveraging a quadratic surrogate modeling approach to replace the location of the overall best with the minimum of a quadratic surrogate as one of the particle attractors.

\emph{Related Literature:}
\ac{PSO} has been shown to successfully optimize a wide range of continuous functions~\cite{KennedyEberhart1995,ClercKennedy2002}, balancing individual and social attributes of particles in the swarm through \textit{exploration} and \textit{exploitation} mechanisms.
The distributed nature of the approach enables rapid and efficient exploration of diverse regions of the design space, increasing the likelihood of discovering high-quality solutions in complex or high-dimensional landscapes~\cite{BonyadiMichalewicz2017}.
Conversely, the exploitation mechanism leverages information gathered from the best-performing solutions to guide particles toward promising regions of the search space, typically using each particle’s personal best and the global best as \textit{attractors}.
While straightforward to implement and commonly adopted, using the overall best as the attractor for the entire swarm can lead to premature convergence to local minima, particularly in functions that pose challenges for hill-climbing.

Hybrid formulations employ surrogate-based optimization algorithms~\cite{WangShanEtAl2004,KazemiWangEtAl2011,QueipoHaftkaEtAl2005}, such as sequential quadratic programming~\cite{QaraadAmjadEtAl2023}, to refine the search around the optimal agent and thereby enhance exploitation capability and solution accuracy~\cite{ZhangWangEtAl2015}.
For very high‑dimensional problems, other authors have employed surrogate‑assisted \ac{PSO} approaches that incorporate radial basis function networks and Gaussian process regression, using information gathered from the swarm’s search process~\cite{TangChenEtAl2013,CaiQiuEtAl2019}.

However, these methods involve cumbersome implementations, which limits their practicality and makes them primarily suitable for large-scale problems.
To address this limitation, a meta‑heuristic is introduced that synthesizes information from multiple distinct best‑performing locations, providing a simple yet robust strategy against local optima while incurring only minimal additional computational cost.
In this context, quadratic models have been applied in areas ranging from large-scale design optimization~\cite{Clemente2025PhD} to control systems~\cite{BemporadMorari1999}, owing to their ability to provide tractable formulations that deliver relatively accurate and computationally efficient solutions~\cite{BoydVandenberghe2004}.
Although numerous \ac{PSO} variants have been developed over the years, to the best of the authors’ knowledge, a meta‑heuristic that leverages quadratic surrogate modeling to identify a better‑conditioned attractor has not been explored.

\emph{Statement of Contribution:}
This paper introduces a \ac{PSO} algorithm that incorporates a quadratic surrogate model to guide the search toward the optimum.
In the proposed method, multiple distinct best-performing locations are fitted with a \textit{n}-dimensional quadratic form, whose properties are exploited to analytically determine its minimum.
During the particle position update phase, this minimum replaces the conventional global-best solution as a dynamic attractor, providing a refined convergence target informed by the local landscape geometry.
The surrogate model is iteratively updated by revising the set of best-performing locations as the swarm explores the design space, progressively refining the approximation of the minimum region.
By exploiting the information already obtained through function evaluations, the method improves the accuracy of the search direction without increasing the number of particles or iterations, which is particularly advantageous for problems with expensive objective evaluations.
Because the surrogate is constructed from multiple best-performing samples, its minimum reflects a consensus of the swarm rather than an individual particle. 
This property reduces the risk of premature convergence to local minima and improves robustness in noisy optimization environments, while introducing only minimal computational overhead compared to conventional \ac{PSO}.

\emph{Organization:}  
The remainder of this paper is organized as follows. Section~\ref{sec:methodology} presents the methodology, highlighting the key aspects of the different algorithms used. Section~\ref{sec:results} compares the method to a standard \ac{PSO} for various benchmark functions to evaluate performance. Finally, Section~\ref{sec:conclusions} concludes the paper and provides an outlook on future research directions.

\section{Methodology}\label{sec:methodology}

In the proposed method, the numerous function evaluations performed by \ac{PSO} are exploited to progressively refine information about the landscape of the objective function.
The number of points required to construct the surrogate function depends on the problem dimensions.
In general, a quadratic function in $n$ variables contains a number of independent coefficients equal to one constant term $c \in \mathbb{R}$, $ n $ linear terms contained in the vector $a \in \mathbb{R}^n$, and $\tfrac{n(n+1)}{2}$ quadratic terms corresponding to the unique squared and cross-product coefficients of the symmetric matrix $B \in \mathbb{R}^{n \times n}$.
Hence, the surrogate model takes the general form 
\begin{equation}\label{eq:quadratic}
	\hat{f}(x) = c + a^\top x + x^\top B x.
\end{equation}
Each sampled point provides one equation constraining these coefficients.
Consequently, at least $ N_{\mathrm{Q}} $ distinct points are required to uniquely determine the quadratic function
\begin{equation}\label{eq:Nq}
	N_{\mathrm{Q}} = 1 + n + \tfrac{n(n+1)}{2} = \tfrac{(n+1)(n+2)}{2}.
\end{equation}
The best $N_{\mathrm{Q}}$ points are determined through a comparison procedure that follows the function evaluation of each particle, described in Algorithm~\ref{alg:optimum}.
Identifying these points requires the same number of function evaluations and only slightly more comparisons than selecting the conventional global best, except in the trivial case where the number of particles equals $N_{\mathrm{Q}}$, in which there is no need for comparisons.
\begin{algorithm}\small
	\caption{Update of the set $\mathcal{Q}$ (best $N_{\mathrm{Q}}$ locations)}
	\label{alg:optimum}
	\KwIn{Current best locations $\mathcal{Q}$, new candidate $X_i^k$, number of stored points $N_{\mathrm{Q}}$}
	\KwOut{Updated best locations $\mathcal{Q}$}
	
	Compute the objective function value $f(x)$\;
		\If{Archive does not exist}{initialize empty heap $H$}

		\If{minimization}{
			\If{$|H| < N_{\mathrm{Q}}$}{ insert $(-f(X_i^k), X_i^k)$ into $H$ }
			\ElseIf{$f(X_i^k)$ improves upon worst in $H$}{replace worst with $(-f(X_i^k), X_i^k)$}
			
		}
		\ElseIf{maximization}{
			\If{$|H| < N_{\mathrm{Q}}$}{ insert $(f(x), x)$ into $H$ }
			\ElseIf{$f(X_i^k)$ improves upon worst in $H$}{replace worst with $(f(X_i^k), X_i^k)$}
			
		}
	\Return{$\mathcal{Q}$}
\end{algorithm}
This comparison is implemented using a heap structure $H$, which enables efficient insertion and replacement of solutions.
Although the order of the points does not affect the interpolation, it is enforced so that, if the interpolation fails due to collinearity, the method falls back to the overall best solution as the attractor.


Once all the particles have been evaluated, the quadratic surrogate model of the objective function is constructed (or updated) using the set of sampled points stored in $\mathcal{Q}$.
Algorithm~\ref{alg:QFND} outlines the procedure for determining the coefficients of the quadratic curve by solving the corresponding linear system and computing the minimizer
\begin{equation}\label{eq:2}
	x_{\min} = -\tfrac{1}{2} B^{-1} a,
\end{equation}
which directly follows from setting the gradient of the quadratic surrogate to zero.
If the points are collinear, resulting in the matrix $B$ being non-invertible, the method falls back to the conventional approach, returning the overall best solution. 
This also occurs when fewer than  $N_{\mathrm{Q}}$ particles are available.
In general, for $N_{\mathrm{p}}$ particles, \ac{PSO} has a computational cost of $O\left(n \, n_{\mathrm{i}} \, N_{\mathrm{p}} \, C_{\mathrm{obj}}\right)$, where $n_{\mathrm{i}}$ is the number of iterations and $C_{\mathrm{obj}}$ is the cost of a single objective function evaluation.
The additional cost introduced by the quadratic fitting step involves solving a dense linear system of size $N_{\mathrm{Q}}$, which has a complexity of $O(N_{\mathrm{Q}}^{3})$~\cite{golub2013matrix}.
Hence, the overall computational cost becomes $O\left(n \, n_{\mathrm{i}}\left(N_{\mathrm{p}} C_{\mathrm{obj}} + N_{\mathrm{Q}}^{3}\right)\right)$.
Consequently, for expensive objective function evaluations such as multi-physics models, for which the method is primarily intended, the additional quadratic-fit overhead remains comparatively small.

\begin{algorithm}\small
	\caption{Quadratic surrogate in $n$ dimensions interpolation and minimization.}
	\label{alg:QFND}
	\KwIn{Set of $N_{\mathrm{Q}}$ points $\mathcal{Q}$, objective function $f$}
	\KwOut{Estimated surrogate minimizer $x_{\min}$, function value $f_{\min}$} %
	
	Initialize coefficient matrix $\mathbf{M} \in \mathbb{R}^{N_{\mathrm{Q}} \times 
	N_{\mathrm{Q}}}$\;
	
	Set column index $c \gets 0$\;
	
	Fill first column of $\mathbf{M}$ with ones (constant term)\;
	Increment $c$\;
	
	\For{$i = 1$ \KwTo $n$}{
		Fill column $c$ of $\mathbf{M}$ with coordinate $i$ of all points in $\mathcal{Q}$ (linear term)\;
		Increment $c$\;
	}
	
	\For{$i = 1$ \KwTo $n$}{
		\For{$j = i$ \KwTo $n$ dimensions}{
			Fill column $c$ of $\mathbf{M}$ with product of coordinates $i$ and $j$ of all points in $\mathcal{Q}$ (quadratic term)\;
			Increment $c$\;
		}
	}
	
	Evaluate objective values $\mathbf{f} = [f(q_1), \dots, f(q_{N_\mathrm{Q}})]^T$\;
	
	Solve interpolation system $\mathbf{M}\theta = \mathbf{f}$\;
	
	Extract linear coefficients $a$ from $\theta$\;
	Construct symmetric quadratic coefficient matrix $B$ from $\theta$\;
	
	\eIf{$B$ is invertible}{
		Compute minimizer $x_{\min} = -\tfrac{1}{2} B^{-1} a$\;
		Compute function actual value $f_{\min}$\;
	}{\tcp{Fallback if $B$ is singular}
		Set $x_{\min} \gets$ first point in $\mathcal{Q}$\;
		Set $f_{\min} \gets$ pre-computed value of $f_{\min}$\;
	}
	
	\Return{$x_{\min}$, $f_{\min}$} %
\end{algorithm}

Once the location of the minimum and its \textit{actual} (non-surrogate) function value $f_{\min}$ are computed, if this value is lower than the overall best found by the particles, it is used as an alternative attractor for the swarm when updating particle velocities, replacing the conventional global best solution
\begin{equation}\label{eq:V_update}
	 V_i^{k+1} = \omega^k V_i^k + c_1^k r_1 (B_i^k - X_i^k) + c_2^k r_2 (x_{\min} - X_i^k),
\end{equation}
and subsequently update the position for the next iteration as
\begin{equation}\label{eq:X_update}
	 X_i^{k+1} = X_i^k + V_i^{k+1}. 
\end{equation}
The coefficients $c_1$ and $c_2$ represent the particles’ individual cognitive and social parameters, respectively, while $r_1$ and $r_2$ are random values drawn from a uniform distribution over [0,1] ($U[0,1]$). These stochastic components enhance the method’s ability to avoid premature convergence to local optima while preserving strong local search capability~\cite{DaiChenEtAl2018}.
By accepting the surrogate minimizer only when it improves upon the current global best, the positive definiteness of $B$ is implicitly enforced, ensuring that the quadratic surrogate is convex rather than concave when minimizing.
Due to the stochastic and derivative‑free nature of \ac{PSO}, establishing global optimality guarantees or classical convergence rates is challenging.
Most theoretical analyses focus on stability properties and conditions under which particles converge to an equilibrium point.
In practice, the quality of the obtained solution depends strongly on the objective function landscape and the algorithm parameters.
To promote exploitation over time and ensure stability, the weights of the attractors are adjusted as the iterations progress~\cite{ShiEberhart1998,SermpinisTheofilatosEtAl2013}.
In particular, the inertia $\omega$ and the individual cognitive parameter $c_1$ decay linearly, while the social parameter $c_2$ increases linearly according to
\begin{equation}\label{eq:omega_update}
	\omega^{k+1} = \omega_{0} - \frac{k}{2 K},
\end{equation}
\begin{equation}\label{eq:c1_update}
	c_1^{k+1} = c_{1,0} - \frac{k}{K},
\end{equation}
\begin{equation}\label{eq:c1_update}
	c_2^{k+1} = c_{2,0} + \frac{k}{K},
\end{equation}
where $k$ denotes the current iteration of the algorithm and $K$
the maximum number of iterations. At the same time, the maximum velocity $v_{\max}$ decays exponentially~\cite{HeMaEtAl2016} as
\begin{equation}\label{eq:c1_update}
	v_{\max}^{k+1} = v_{\max,0} \cdot e^{\left(1-\frac{k}{K}\right)}.
\end{equation}
However, to counteract the risk that the decay of the inertia parameter may cause particles to become trapped in local minima, the exploration safeguard described in Section~III-C of Kwok et al.~\cite{KwokHaEtAl2007} is implemented.
Specifically, the current value of a particle is compared with its position observed $S$ iterations earlier.
If the two values are too similar ($\gamma_i^k  < 0.5$), the particle’s inertia is multiplied by the coefficient $\tau$, as expressed in the following equations
\begin{equation}\label{eq:expsafeguard}
	\gamma_i^k = \frac{\mid f(X_i^k) - f(X_i^{k-S}) \mid}{\mid f(X_i^{k-S})\mid},
\end{equation}
\begin{equation}\label{eq:omegatau}
	\omega_i^k = \omega_i^k \tau.
\end{equation}
A small lower bound is imposed on $\gamma$ to avoid numerical ill-conditioning for values that approach zero.
Finally, if the new position of a particle lies outside the optimization boundaries, the position is clipped.
Similarly, if the velocity exceeds the maximum velocity $v_{\max}$, it is also clipped.



\begin{figure*}
	\begin{minipage}[b]{.48\linewidth}
		\centering
		\centerline{\includegraphics[width=0.75\linewidth]{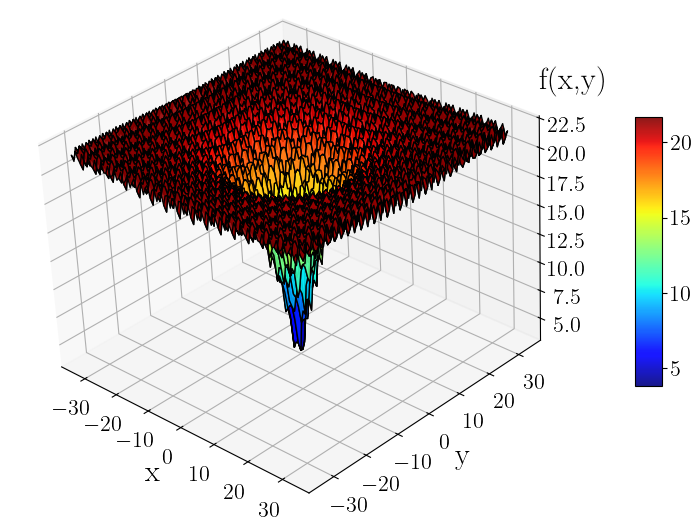}}
	\end{minipage}
	\begin{minipage}[b]{.48\linewidth}
		\centering
		\centerline{\includegraphics[width=0.8\linewidth]{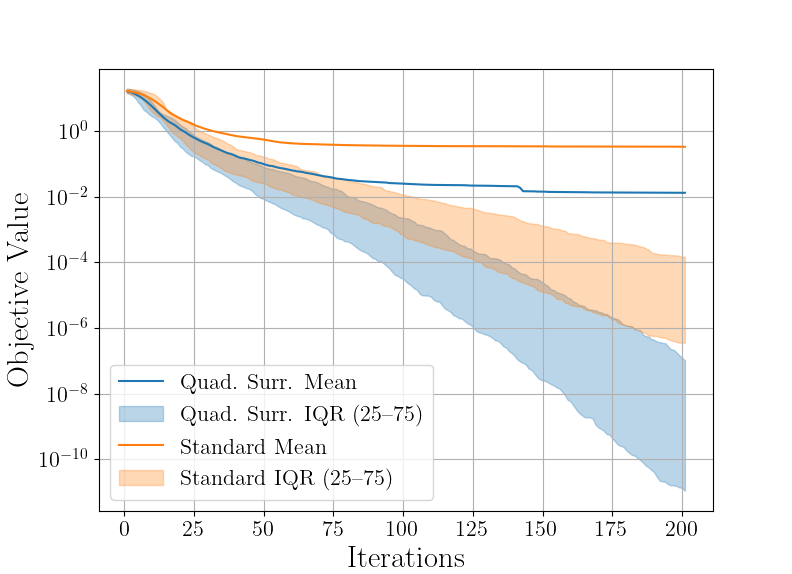}}
	\end{minipage}
	\caption{Ackley’s function is shown in the left panel. The right panel compares the quadratic surrogate attractor with the standard algorithm over 400 independent runs of 200 iterations each. The solid line represents the mean across runs, while the shaded area denotes the inter-quartile range (25–75\%). }
	
	\label{fig:Ackley}
\end{figure*}

\begin{figure*}
	\begin{minipage}[b]{.48\linewidth}
		\centering
		\centerline{\includegraphics[width=0.7\linewidth]{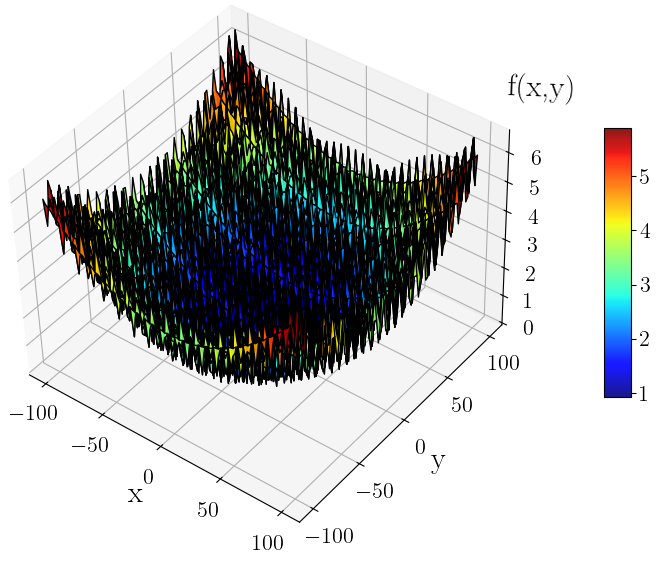}}
	\end{minipage}
	\begin{minipage}[b]{.48\linewidth}
		\centering
		\centerline{\includegraphics[width=0.8\linewidth]{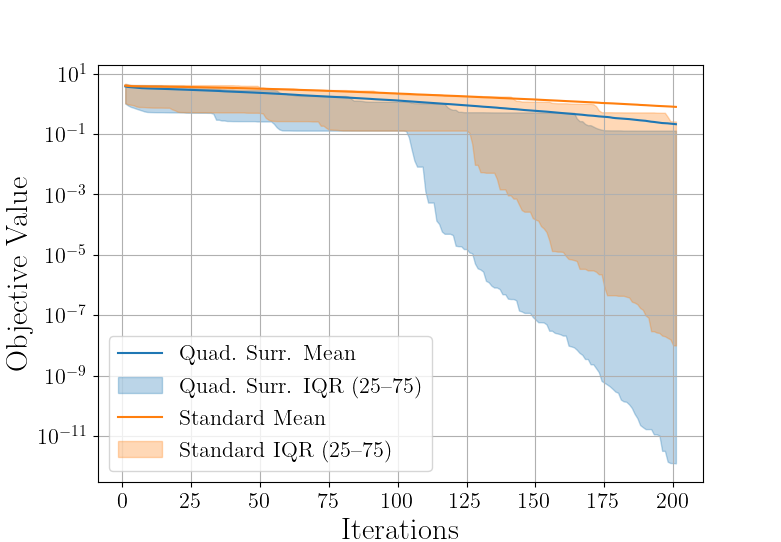}}
	\end{minipage}
	\caption{Griewank’s function is shown in the left panel. The right panel compares the quadratic surrogate attractor with the standard algorithm over 400 independent runs of 200 iterations each. The solid line represents the mean across runs, while the shaded area denotes the inter-quartile range (25–75\%).}
	\label{fig:Griewank}
\end{figure*}

\section{Results}\label{sec:results}

In this section, the effectiveness of the proposed algorithm is demonstrated through a numerical study on a set of benchmark optimization functions~\cite{VirtualLibrary}.
These functions exhibit diverse landscapes and impose varying levels of difficulty for iterative optimizers.
A total of 400 independent test runs of 200 iterations each are conducted for every function–algorithm pair, and the results are evaluated based on their statistical characteristics, with the merits assessed through the corresponding result distributions.
The performance of the quadratic surrogate minimum attractor algorithm is compared with that of a standard \ac{PSO}, using identical parameter settings as reported in Table~\ref{tab:settings}.
All optimizations have been performed on a laptop equipped with an Intel(R) Core(TM) i7-9750H CPU @ 2.60 GHz, without the use of parallel computing.
The full implementation of the algorithm has been released as open‑source code and can be accessed at: \mbox{\url{https://github.com/YAPSO-team/YAPSO}}.\\

\begin{figure*}
	\begin{minipage}[b]{.33\linewidth}
		\centering
		\centerline{\includegraphics[width=0.9\linewidth]{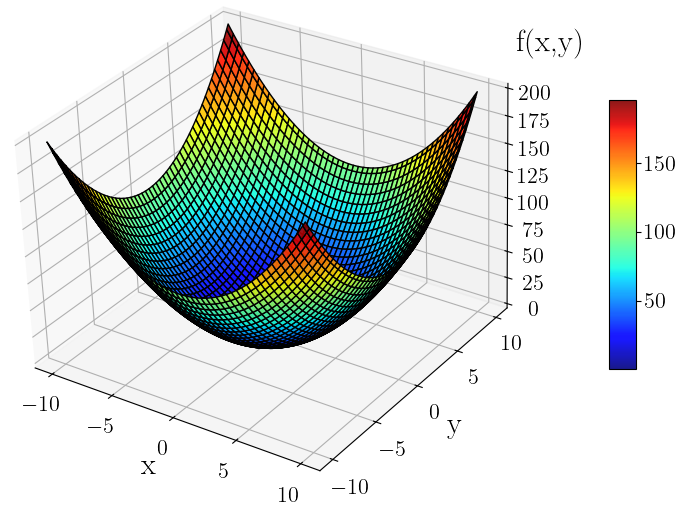}}
		\small\centerline{Sphere function}\medskip
	\end{minipage}
	\begin{minipage}[b]{.33\linewidth}
		\centering
		\centerline{\includegraphics[width=\linewidth]{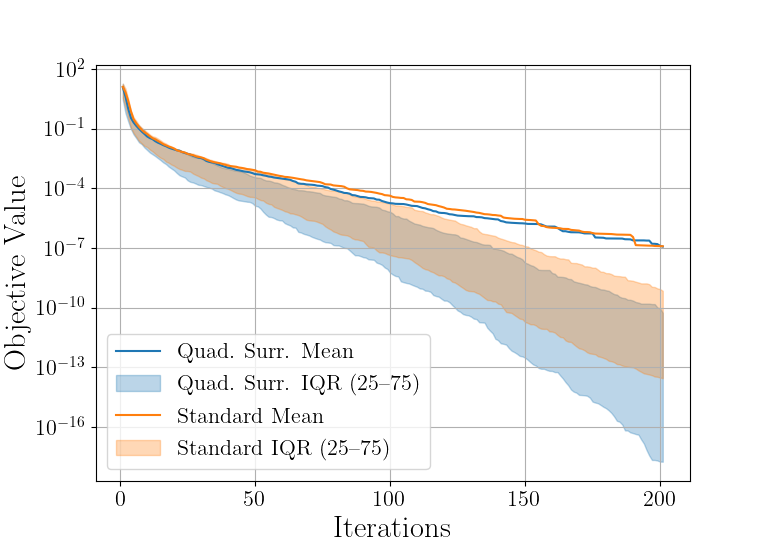}}
		\small\centerline{2D Sphere function, 6 particles}\medskip
	\end{minipage}
	\begin{minipage}[b]{.33\linewidth}
		\centering
		\centerline{\includegraphics[width=\linewidth]{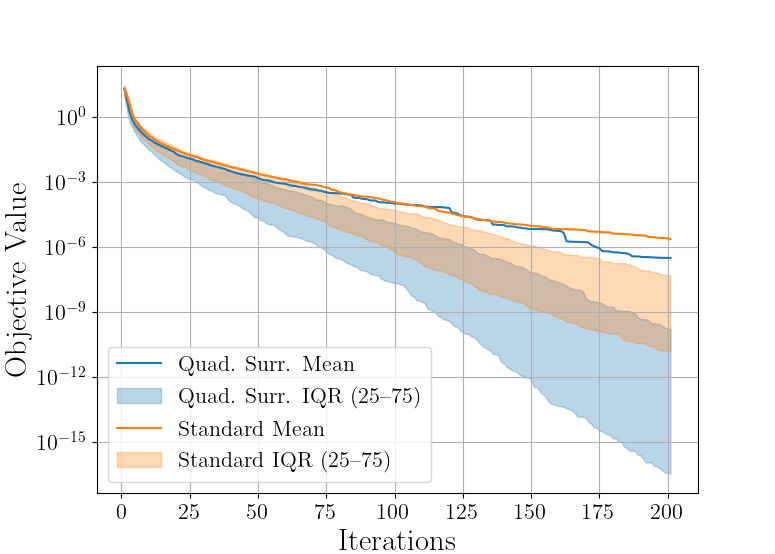}}
		\small\centerline{3D Sphere function, 10 particles}\medskip
	\end{minipage}
	\caption{The Sphere function is shown in the left panel. The central panel illustrates algorithms performance for the two‑dimensional case with six particles, while the right panel presents the three‑dimensional case with ten particles. In both cases, the quadratic surrogate attractor is compared with the standard algorithm over 400 independent runs of 200 iterations each. The solid line represents the mean across runs, while the shaded area denotes the inter-quartile range (25–75\%).}\label{fig:Sphere}
\end{figure*}

\begin{figure*}
	\begin{minipage}[b]{.33\linewidth}
		\centering
		\centerline{\includegraphics[width=0.9\linewidth]{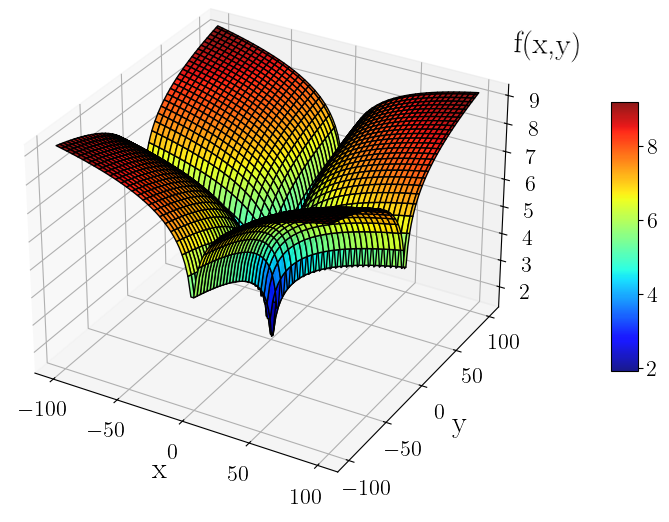}}
		\small\centerline{\textit{Flower} function}\medskip
	\end{minipage}
	\begin{minipage}[b]{.33\linewidth}
		\centering
		\centerline{\includegraphics[width=\linewidth]{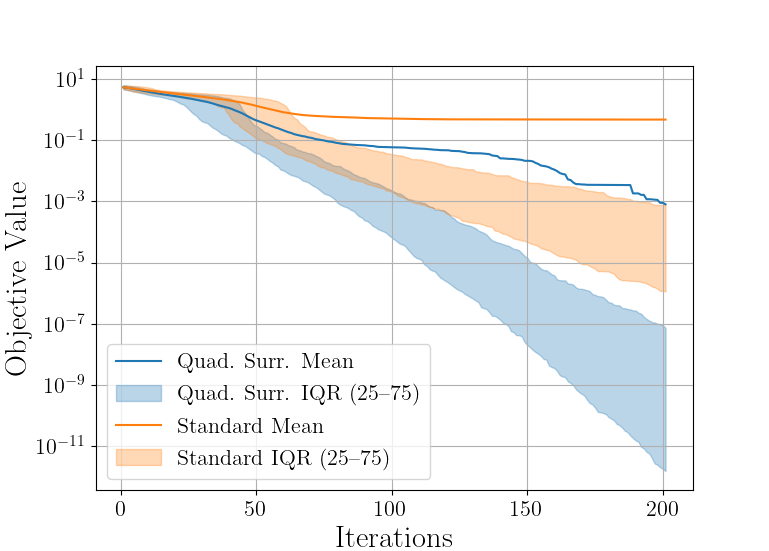}}
		\small\centerline{2D \textit{Flower} function, 6 particles}\medskip
	\end{minipage}
	\begin{minipage}[b]{.33\linewidth}
		\centering
		\centerline{\includegraphics[width=\linewidth]{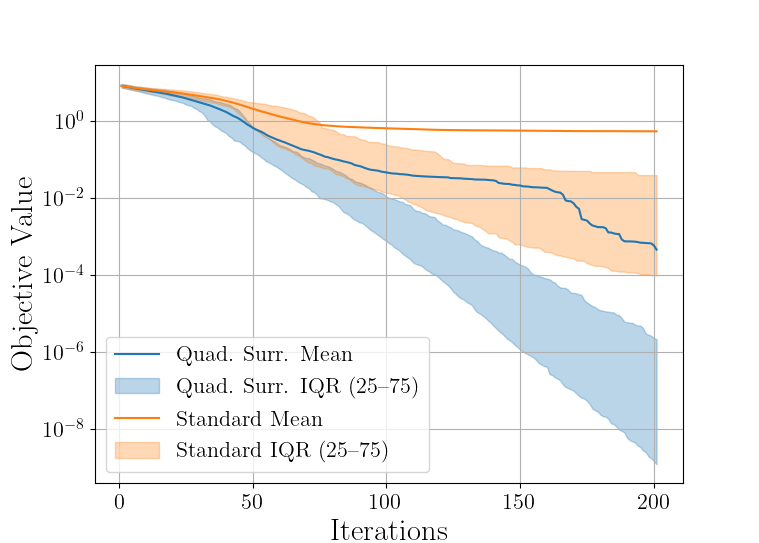}}
		\small\centerline{3D \textit{Flower} function, 10 particles}\medskip
	\end{minipage}
	\caption{The \textit{Flower} function is shown in the left panel. The central panel illustrates algorithms performance for the two‑dimensional case with six particles, while the right panel presents the three‑dimensional case with ten particles. In both cases, the quadratic surrogate attractor is compared with the standard algorithm over 400 independent runs of 200 iterations each. The solid line represents the mean across runs, while the shaded area denotes the inter-quartile range (25–75\%).}\label{fig:Flower}
\end{figure*}

First, the two methods are applied to Ackley’s~\cite{Ackley1987} and Griewank’s~\cite{Griewank1981} functions, shown in Figures~\ref{fig:Ackley} and \ref{fig:Griewank}, respectively.
Ackley’s function~\cite{Ackley1987} shows a large number of local minima surrounding the global minimum at $(0,0)$. Particles can easily become trapped in these local minima, highlighting the function’s multimodal and challenging nature for global search methods.
Griewank’s function~\cite{Griewank1981} is characterized by a combination of a broad parabolic shape and superimposed cosine oscillations. Its regularly spaced local minima form a challenging but structured landscape for optimization algorithms searching for the global minimum at $(0,0)$.
It is observed that the quadratic surrogate attractor approach performs better in both cases, demonstrating greater robustness in handling noisy landscapes and thereby addressing a critical challenge in these algorithms.
The numerical performance of the algorithms, across different functions and particle counts, is reported in Table~\ref{tab:results}.
In both figures, the left panel portrays the function, while the right panel displays performance in terms of the solution accuracy achieved with the number of iterations. 
Due to the logarithmic representation, the mean value may lie outside the inter-quartile range.
\begin{table}[h!]
	\centering
	\caption{Weights and parameters used~\cite{ClercKennedy2002,HeMaEtAl2016,KwokHaEtAl2007}.} 
	\label{tab:settings}
	\begin{tabular}{c c c c c c c}\toprule
		$\mathbf{\omega}$ & $\mathbf{c_{1,0}}$ & $\mathbf{c_{2,0}}$ & $\mathbf{v_{\max,0}}$ & $\mathbf{K}$ & $\mathbf{S}$ & $\mathbf{\tau}$\\ 
		\midrule
		0.72984 & 2.8 & 2.05 & 2 & 200 & 52 & 1.2\\
		\bottomrule
	\end{tabular}
\end{table}

\begin{table*}[t]
	\centering
	\caption{Comparison of the quadratic surrogate attractor (Q.S.) and the standard \ac{PSO} (Std.) across benchmark functions. 
		Values report the mean, execution time, and inter-quartile ranges (IQR 25–75) over 400 independent runs of both algorithms for 200 iterations.
		Negative percentages denote a more accurate solution.}
	\label{tab:results}
	\resizebox{\textwidth}{!}{
		\begin{tabular}{l c c | c c c | c c c|c c | }\toprule
			\textbf{Function} & \textbf{N$_\mathrm{p}$} & \textbf{Bounds} & \textbf{Mean Q.S.} &\textbf{Mean Std.} & \textbf{Rel. Diff.} & \textbf{Time Q.S. [\unit{s}]} & \textbf{Time Std. [\unit{s}]} & \textbf{Rel. Diff.} & \textbf{IQR 25-75 Q.S.} & \textbf{IQR 25-75 Std.} \\ 
			\midrule
			Ackley 2D & 6 & [[-32.768, 32.768],[-32.768, 32.768]] & 1.307e-2 & 3.294e-1 & -96.03\% & 91.27 & 84.41 & + 8.12 \% & (1.112e-11)–(3.568e-7) & (1.037e-7)–(1.478e-4) \\ 
			
			Griewank 2D & 6 & [[-600, 600],[-600, 600]] & 2.119e-1 & 7.927e-1 & -73.27\% & 117.55 & 102.41 & + 14.78 \% &  (1.265e-12)–(1.300e-1) & (1.010e-8)–(2.610e-1) \\
			
			Sphere 2D & 6 & [[-10, 10],[-10, 10]] & 1.182e-7 & 1.235e-7 & -4.292\% & 67.19 & 60.32 & + 11.38 \% &  (1.791e-18)–(5.523e-11) & (2.838e-14)–(7.071e-10) \\
			
			Sphere 3D & 10 & [[-10, 10],[-10, 10],[-10, 10]] & 3.122e-7 & 2.334e-6 & -86.62\% & 75.09 & 67.59 & + 11.10 \% &  (3.306e-17)–(1.637e-10) & (1.525e-11)–(4.652e-8) \\
			
			Flower 2D & 6 & [[-100, 100],[-100, 100]] & 8.045e-4 & 4.776e-1 & -99.83\% & 76.43 & 69.21 & + 10.43 \% &  (1.501e-12)–(7.327e-8) & (1.132e-6)–(7.560e-4) \\ 
			
			Flower 3D & 10 & [[-100, 100],[-100, 100],[-100, 100]] & 4.565e-4 & 5.374e-1 & -99.92\% & 96.25 & 87.39 & + 10.14 \% &  (1.221e-9)–(2.168e-6) & (9.853e-5)–(3.932e-2) \\
			\bottomrule
		\end{tabular}
	}
\end{table*}
\vspace{10pt}
Figures~\ref{fig:Sphere} and \ref{fig:Flower} analyze the performance of the quadratic surrogate attractor approach in comparison with the standard \ac{PSO} algorithm for the sphere function
\begin{equation}\label{eq:sphere}
	f(\mathbf{x}) = \sum_{i=1}^{n} x_i^2,
\end{equation}
and the \textit{Flower} function,
\begin{equation}\label{eq:flower}
f(\mathbf{x}) = \sum_{i=1}^{n} \log\!\bigl(|x_i| + 1\bigr),
\end{equation}
a more challenging, quasi-convex function.
The sphere function is characterized by a simple convex quadratic shape centered at the origin. Its smooth landscape contains a single global minimum at $(0,0)$, making it a standard benchmark for evaluating the convergence speed and accuracy of optimization algorithms.
The flower function is characterized by logarithmic growth along both coordinate axes, producing a landscape with a sharp basin near the origin and gradually flattening slopes outward. It also has a single global minimum at $(0,0)$, while the nonlinear curvature provides a distinct challenge compared to purely quadratic functions.
To analyze the effect of dimensionality on the method, the two‑dimensional and three‑dimensional cases of these functions are considered.
In the 3D case, ten particles are used, which is the minimum required to construct a three‑dimensional surrogate.
It is observed that more complex functions benefit from this augmentation, whereas simpler functions (such as the 2D Sphere) only show little improvement.
In line with the results obtained from other, more computationally demanding surrogate-assisted approaches~\cite{CaiQiuEtAl2019}, the proposed quadratic surrogate attractor \ac{PSO} algorithm consistently outperforms the standard \ac{PSO} in terms of solution quality for a fixed number of iterations.
The improvement is particularly pronounced in the case of quasi-convex functions, where the surrogate model is able to exploit the underlying convex-like structure of the landscape.
%
%
Given enough particles to build the surrogate ($N_{\mathrm{p}} \geq N_{\mathrm{Q}}$), the approach is well suited even for higher-order functions, since any interior minimizer behaves approximately quadratically locally, and the surrogate captures the predominant second‑order behavior around the minimizer.
Additionally, any objective function whose minimum lies on the boundary of the feasible set results in the surrogate minimizer lying either on the boundary or outside the feasible region, thereby directing the particles toward the boundary in both cases.
Since the surrogate is quadratic and its gradient is analytically available, the constrained minimizer under box constraints could in principle be computed directly via the Karush–Kuhn–Tucker conditions. 
This capability is not exploited in the present method, but it suggests a potential extension for a more rigorous treatment of boundary optima than relying on the unconstrained minimizer falling outside the feasible region.
Finally, the effect of extending the velocity update rule by incorporating the surrogate-based attractor alongside the conventional global best is investigated. This modification degraded performance, likely due to conflicting attractor dynamics that reduced the swarm’s ability to converge effectively. Replacing the global best with the surrogate-based attractor, rather than combining them, therefore appears to be the more effective strategy.




\section{Conclusions}\label{sec:conclusions}

This paper proposed a quadratic surrogate modeling approach to augment the \ac{PSO} algorithm by replacing the overall best solution in favor of a better‑conditioned dynamic attractor for the swarm.
The refined convergence target, informed by the local landscape, is determined by interpolating multiple distinct best‑performing locations and analytically computing the minimum of the quadratic function.
This method improves resilience against premature convergence and noisy environments while maintaining minimal computational overhead.
To ensure statistical significance, 400 independent runs of 200 iterations were conducted for each function–algorithm pair, analyzing statistical characteristics and corresponding distributions.
The surrogate attractor approach demonstrated consistently higher accuracy than the standard algorithm in all cases considered, with only a modest increase in computational time.

These findings open the way for future extensions, including investigations on the impact of limited iterations or termination conditions, a thorough study of the effect of swarm size, and comparisons with other advanced optimization algorithms.


%


\small
%
\bibliographystyle{ieeetr}
\bibliography{ACC}
\raggedbottom

\end{document}